%% file: v1_mikhail.tex
\documentclass[10pt,twocolumn,letterpaper]{article}

\usepackage{cvpr}              

\input{preamble}

\definecolor{cvprblue}{rgb}{0.21,0.49,0.74}
\usepackage[pagebackref,breaklinks,colorlinks,citecolor=cvprblue]{hyperref}

\usepackage[accsupp]{axessibility} 

\usepackage{amsmath}
\usepackage{algorithm}
\usepackage{algorithmic}
\usepackage{tabularx,ragged2e}
\usepackage{pifont}
\newcommand{\cmark}{\ding{51}}%
\newcommand{\xmark}{\ding{55}}%

\newcolumntype{L}{>{\centering\arraybackslash}X}
\newcommand{\PreserveBackslash}[1]{\let\temp=\\#1\let\\=\temp}
\newcolumntype{C}[1]{>{\PreserveBackslash\centering}p{#1}}
\newcommand{\midline}{\noalign{\vskip 0.01cm} \hline \noalign{\vskip 0.01cm}}
\newcommand{\topline}{\hline \noalign{\vskip 0.01cm}}

\usepackage{color,soul}
\usepackage{bbding}
\usepackage{ulem} 
\soulregister\cite7
\soulregister\ref7
\soulregister\pageref7

\makeatletter
\robustify\@latex@warning@no@line
\makeatother
\usepackage{authblk}

\def\paperID{554} 
\def\confName{CVPR}
\def\confYear{2024}

\title{CAT: Exploiting Inter-Class Dynamics for Domain Adaptive Object Detection}

\author[1,2]{Mikhail Kennerley}
\author[2]{Jian-Gang Wang}
\author[1]{Bharadwaj Veeravalli}
\author[3,1]{Robby T. Tan}
\affil[1]{National University of Singapore, Department of Electrical and Computer Engineering} 
\affil[2]{Institute for Infocomm Research, A*STAR} %
\affil[3]{ASUS Intelligent Cloud Services}
\affil[ ]{\tt\small mikhailk@u.nus.edu, jgwang@i2r.a-star.edu.sg, elebv@nus.edu.sg, robby.tan@nus.edu.sg}

\begin{document}
\maketitle


\begin{abstract}
Domain adaptive object detection aims to adapt detection models to domains where annotated data is unavailable.
Existing methods have been proposed to address the domain gap using the semi-supervised student-teacher framework.
However, a fundamental issue arises from the class imbalance in the labelled training set, which can result in inaccurate pseudo-labels.
The relationship between classes, especially where one class is a majority and the other minority, has a large impact on class bias.
We propose Class-Aware Teacher (CAT) to address the class bias issue in the domain adaptation setting.
In our work, we approximate the class relationships with our Inter-Class Relation module (ICRm) and exploit it to reduce the bias within the model.
In this way, we are able to apply augmentations to highly related classes, both inter- and intra-domain, to boost the performance of minority classes while having minimal impact on majority classes.
We further reduce the bias by implementing a class-relation weight to our classification loss. 
Experiments conducted on various datasets and ablation studies show that our method is able to address the class bias in the domain adaptation setting.
On the Cityscapes $\to$ Foggy~Cityscapes dataset, we attained a 52.5 mAP, a substantial improvement over the 51.2 mAP achieved by the state-of-the-art method. \footnote{\url{www.github.com/mecarill/cat}}
\end{abstract}


\section{Introduction}
\label{sec:intro}

\begin{figure}[t]
\includegraphics[width=0.5\textwidth]{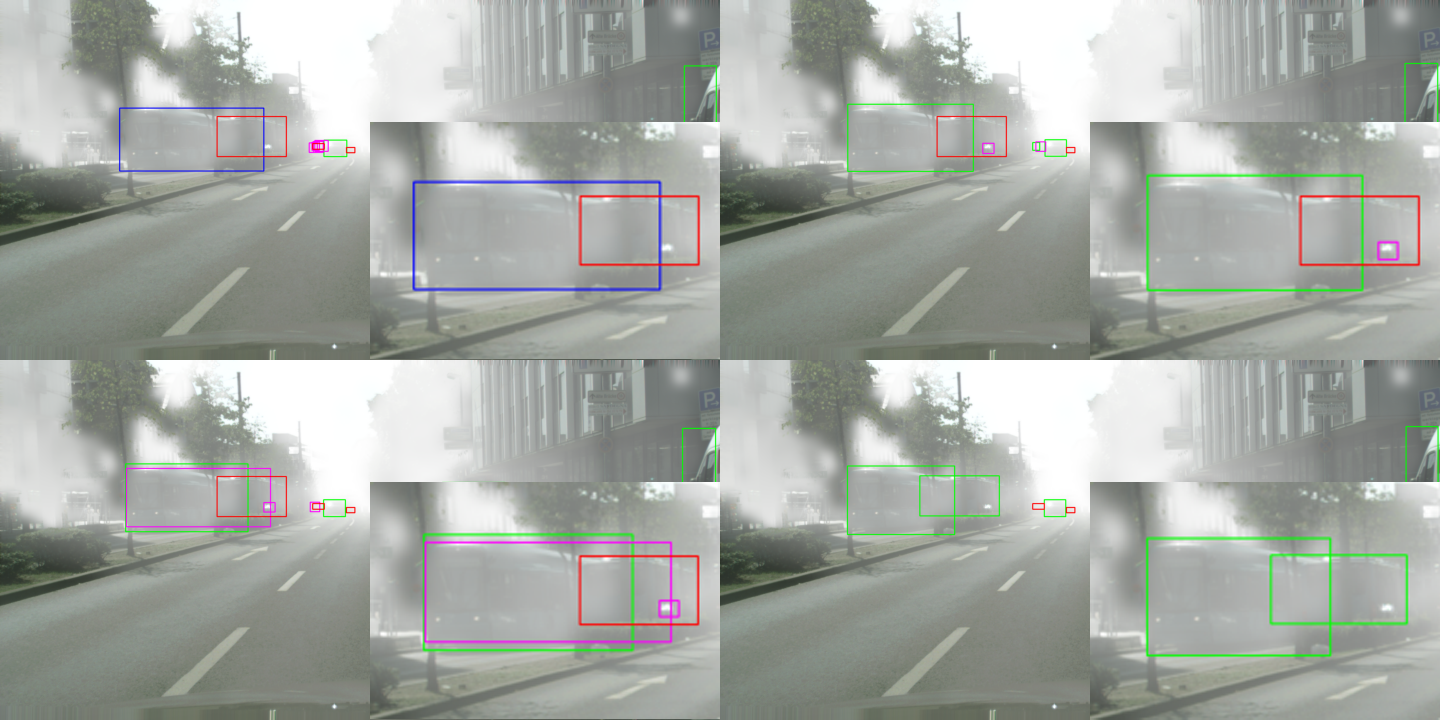}
\caption{\textbf{Performance of Class-Aware Teacher (CAT).} AT \cite{at} (top left), with Inter-Class Loss, ICL, (top-right), with Class Relation Augmentation, CRA, (bottom-left), and CAT (bottom-right). CAT is able to address misclassification and false positives, blue and red boxes, respectively, in minority classes such as 'train'. The combination of ICL and CRA further boosts performance by reducing the number of false positives shown as pink boxes.}
\centering
\label{fig:teaser}
\end{figure}

Domain adaptive object detection (DAOD) has been proposed as a solution for object detection in domains where no annotated data is available.
This need is due to the increasing demands of data tied with annotation being both cost-prohibitive and potentially inaccurate in challenging domains.
DAOD has been progressing with the introduction of adversarial learning \cite{dafaster,strongweak,gpa,tia}, style transfer \cite{cyclegan,forkgan}, and notably, student-teacher frameworks \cite{meanteacher,umt,at,tdd,cmt,2pcnet,mila,mrt}.
Yet, these methods ignore the critical issue of class imbalance, which is a problem in many real-life scenarios, such as autonomous driving. 
For instance, in the Cityscapes  dataset~\cite{cityscapes}, the 'car' class dominates the dataset with 26,963 instances while classes such as 'train' contain only 168 instances.

Previous work to mitigate class imbalance in DAOD has applied class-specific discriminators \cite{tia} to align classes in distinct domains. 
Additionally, class weights has been proposed to boost minority categories while aligning domain features \cite{i3net}.
Recently, many DAOD methods employ the student-teacher framework, leading to improved performance. 
Despite their effectiveness, these student-teacher methods suffer from the class imbalance problem, resulting in poor performance for minority classes.

In the student-teacher framework, class-specific thresholding, which provides more lenient thresholds for minority classes, has been proposed~\cite{Li2021RethinkingPL,Kar_2023_CVPR,Zhang_Pan_Wang_2022}. 
Yet, this approach does not address the fundamental class imbalance. 
Even with perfectly accurate pseudo-labels guiding the student, the model's bias would at best align with the biases present in the dataset, rather than providing an unbiased view. 
Moreover, inter-class dynamics play a crucial role in addressing class imbalance, especially when minority classes share high similarities with majority classes, increasing the likelihood of misclassification.

To address these challenges, in this paper, we introduce our Class-Aware Teacher (CAT), specifically designed to tackle class imbalance in the DAOD setting.
CAT implements an Inter-Class Relation module (ICRm) that approximates the model's existing class biases as well as inter-class dynamics. 
With the knowledge of these biases, CAT applies Class-Relation Augmentation (CRA) to the training images.
CRA increases the representation of minority classes by blending them with highly similar majority classes at the instance-level.
To aid in this augmentation, a Cropbank \cite{Zhang_Pan_Wang_2022} is used to store a collection of cropped instances.
Furthermore, this augmentation is not just confined to the source domain but is also applied across domains.
By allowing cross-domain augmentation, we are able to address the domain gap more holistically.
To further address the inter-class bias, we propose an Inter-Class Loss (ICL).
ICL utilises the insights from the ICRm to prioritise the model's attention towards minority classes.
This priority is particularly focused on cases where minority classes are prone to being misclassified as majority classes.

By integrating these methods, our results indicate an improvement in the accuracy of minority class predictions, with a quantifiable increase in performance on benchmarks such as Cityscapes $\to$ Foggy Cityscapes by +1.3 mAP.
Figure~\ref{fig:teaser} demonstrates the performance of our method.
We summarise the contributions of this paper as follows:
\begin{itemize}
  \item We propose our Class-Aware Teacher (CAT) model, supported by our inter-class relation module (ICRm), which is able to map the model's existing class biases.
  \item We present Class-Relation Augmentation which emphasises augmentation between related classes across domains, coupled with Inter-Class Loss to further prioritise the performance of minority classes.
  \item Thorough experimental analysis that confirms the capabilities of CAT. Our experiments demonstrate significant improvements in performance compared with the state-of-the-art methods in DAOD benchmarks. 
\end{itemize}

\section{Previous Work}
\label{sec:prevwork}

\paragraph{UDA for Object Detection}
Unsupervised Domain Adaptation (UDA) is designed to adapt a model trained on a labelled source domain to an unlabelled target domain.
In object detection tasks, methods like adversarial training coupled with domain classifiers \cite{dafaster,strongweak,gpa,tia} are prevalent for cultivating domain-invariant image feature representations.
Other strategies, such as image-to-image translation, use generative models \cite{cyclegan, forkgan} or frequency-based methods \cite{fda} to bridge the gap between domains.
Recent approaches have applied the mean-teacher (MT) framework \cite{meanteacher,umt,at,tdd,cmt,2pcnet,mila}, initially conceived for semi-supervised learning, to UDA challenges.
For instance, the UMT \cite{umt} leverages CycleGAN-generated images to train the student-teacher model, aiming to diminish domain bias.
AT \cite{at} employs strong-weak image augmentation, intentionally degrading the student's input compared to the teacher's, and incorporates adversarial training to further reduce the domain gap.
2PCNet \cite{2pcnet} takes a two-stage approach to provide more diverse pseudo-labels with domain specific augmentation.
Despite significant improvements over their predecessors, these methods often overlook the class imbalance issue prevalent in benchmark datasets. 
This oversight can lead to suboptimal performance on minority classes, some of which may appear up to 20 times less frequently than majority classes \cite{cityscapes}.

\paragraph{Class-Imbalanced Object Detection}
The issue of imbalance in object detection largely stems from an overrepresentation of background over foreground classes in predictions \cite{imbalance}. 
Our research, however, addresses the imbalance among foreground classes themselves, which often suffer from unequal frequency within datasets. 
A challenge here is the risk of overfitting to minority classes, particularly when their instances are sparse \cite{he2009learning}.
Over-sampling of minority classes, a strategy borrowed from classification tasks, has been adapted for object detection as well \cite{Zhang_Pan_Wang_2022}. 
In student-teacher frameworks, a static hard threshold for pseudo-label generation has evolved into a dynamic, class-specific threshold to mitigate teacher bias \cite{Li2021RethinkingPL, Kar_2023_CVPR}. 
Although this approach can enhance the quality of pseudo-labels, it does not necessarily balance the sample distribution between majority and minority classes.
In the DAOD, methods such as class-specific discriminators \cite{tia} and weighted losses \cite{i3net} have been proposed to address class imbalance alongside domain adaptation. 
A critical aspect that remains underexplored is the inter-class relationship, particularly between majority and minority classes with similar features.
In this paper, we aim to explore inter-class dynamics to effectively tackle class imbalance.

\begin{figure*}[t]
\centering
\includegraphics[width=0.96\textwidth]{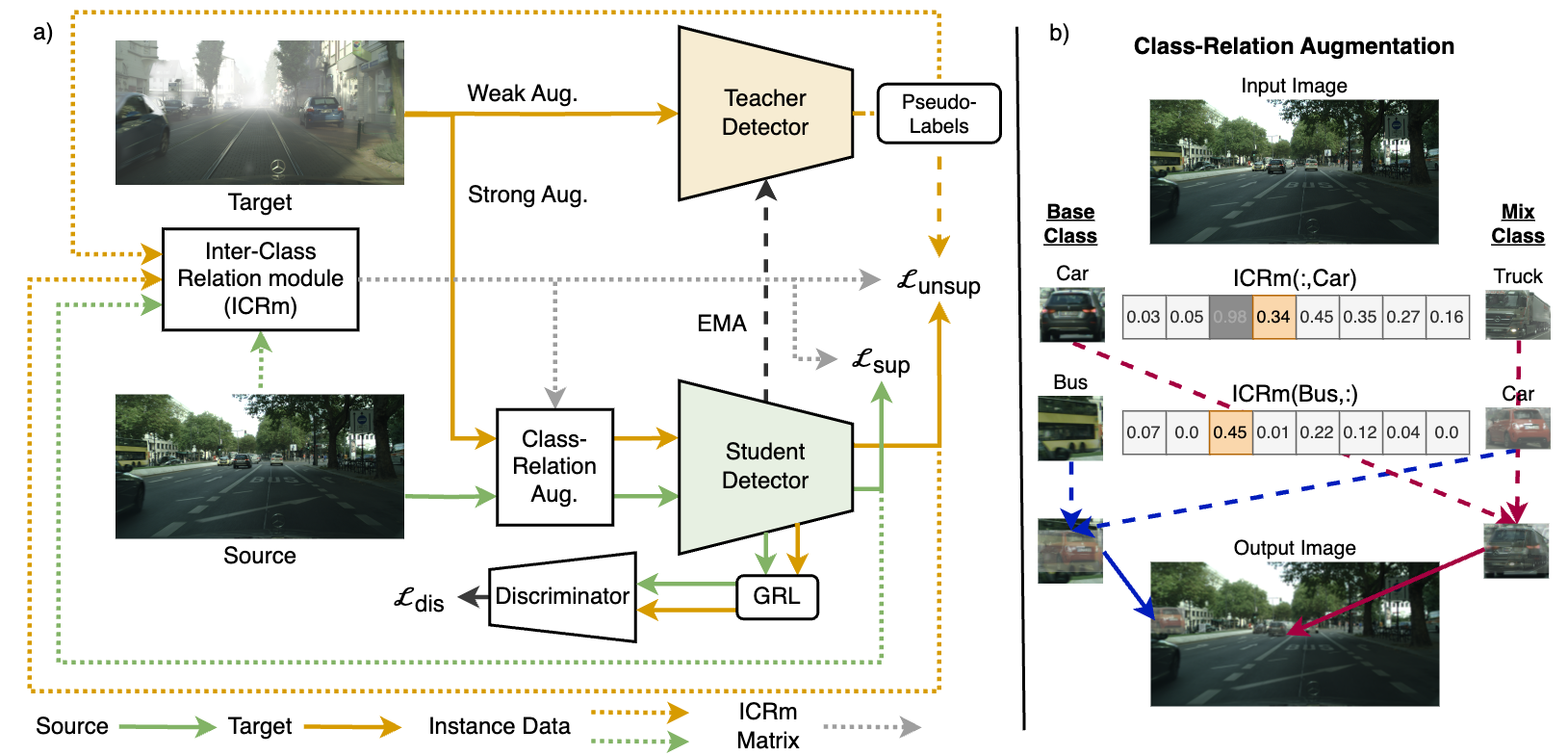}
\caption{(a) \textbf{Class-Aware Teacher (CAT)} consists of: a student-teacher network; Inter-Class Relation module (ICRm), which estimates inter-class biases; Class-Relation Augmentation, which augments images to reduce the inter-class biases by mixing the cropped instances of related classes; and Inter-Class Loss, which emphasises the loss on highly misclassified minority classes. (b) Class-Relation Augmentation demonstrated on majority (Car) and minority (Bus) classes.}
\label{fig:arch}
\end{figure*}

\section{Preliminaries}

\paragraph{Problem Definition}
In this paper, we propose a method for class balanced domain adaptive object detection, employing a labelled source dataset $D_s=\{I_{s};Y_{s}\}$ and an unlabelled target dataset  $D_t=\{I_{t}\}$. 
$I_s$ and $Y_s=\{b_s;c_s\}$ denote the images and their corresponding ground-truth labels, which include bounding box and class information, indicated as 
$b_s$ and $c_s$, respectively. 

\paragraph{Mean Teacher}
\label{sec:MTB}
We utilise the mean-teacher framework, comprising of a student and teacher network that shares identical architectures and network parameters.
The teacher's network parameters, denoted as $\theta_t$, are not updated through backpropagation but are instead updated using the Exponential Moving Average (EMA) of the student's parameters $\theta_s$, following:
\begin{equation}
\theta_t \gets \alpha\theta_t + (1-\alpha)\theta_s,
\end{equation} 
where $\alpha$ is the decay rate that controls the update momentum.

The teacher network generates pseudo-labels , $Y_t=\{b_t,c_t\}$ from weakly augmented unlabelled images.
These pseudo-labels are utilised by the student network to calculate the unsupervised loss in conjunction with the strongly augmented inputs.
The student's inputs are intentionally degraded compared to the teacher's inputs to challenge the student network further.
The supervised loss is consistent with the Faster R-CNN framework \cite{frcnn}, while the unsupervised loss is formulated as:
\begin{equation}
\mathcal{L}_{\rm unsup} =\mathcal{L}_{\rm obj}^{\rm rpn}(I_t,b_t) + \mathcal{L}_{\rm cls}^{\rm roi}(I_t,b_t,c_t).
\end{equation} 
notably excluding the regression losses in the unsupervised context. 
We implement a hard threshold, $\tau$, on the classification scores generated by the teacher to ensure that only pseudo-labels with high confidence are utilised by the student network, thereby encouraging more reliable learning outcomes. 
Following \cite{at}, a discriminator is added to encourage domain invariant feature representations with an associated loss, $\mathcal{L}_{\rm dis}$.

\section{Proposed Method}
\label{sec:method}

Our method, Class-Aware Teacher (CAT), as depicted in Figure~\ref{fig:arch}.a, builds on the mean-teacher framework.
Central to CAT is our Inter-Class Relation module (ICRm), designed to quantify the class biases inherent in the model. 
Unlike traditional approaches that address class bias in a broad sense, ICRm maps the dynamic relationships between classes.
It particularly focuses on minority classes that are disproportionately misclassified as dominant majority classes. 
This mapping is achieved by constructing a confusion matrix for each batch, which is normalised against the ground truth, allowing for real-time bias estimation.
A global matrix, updated continually with batch-level data, serves as a robust representation of the model's class biases.
The ICRm is integral to our methodology, underpinning two components: the Class-Relation Augmentation and the Inter-Class Loss.

Class-Relation Augmentation addresses class imbalance at the image-level. 
Minority classes that share a high similarity to majority classes in an image are identified using ICRm.
MixUp \cite{mixup}, which has been shown to address the imbalanced class problem \cite{balmixup,remix}, is then used to merge the related minority and majority classes thus increasing the representation of these minority classes. 
Our process not only boosts the number of minority class samples but also encourages the model to distinguish between closely associated classes, as illustrated in Figure \ref{fig:arch}.b.

Additionally, ICRm informs the distribution of weights within our Inter-Class Loss.
This weighted loss function emphasises the loss on minority classes, especially those frequently mislabelled as majority classes.
By doing so, we provide a counterbalance to the model's learned biases, nudging it towards a more balanced classification performance.

\subsection{Inter-Class Relation module}
Prior research addressing class imbalance within domain adaptation \cite{hsuUnsupervisedDomainAdaptation2015, jiangImplicitClassConditionedDomain2020, tanwisuthPrototypeOrientedFrameworkUnsupervised2021} has significantly advanced the performance on imbalanced classes. 
However, these methods often overlook the inter-class relationships and their impact on class imbalance. 
Our experimental observations suggest that the likelihood of misclassification between minority and majority classes is heavily influenced by their similarity.
For instance, minority vehicle classes are more prone to be misclassified as 'car', a majority class, rather than as 'person', another majority class, due to their inherent resemblance.

Our approach aims to leverage these observed relationships through the Inter-Class Relation module (ICRm). 
Distinct from general class bias, inter-class dynamics cannot be directly inferred from the dataset but must be extrapolated from the model during training.
We achieve this by generating a confusion matrix at each training batch that cross-references the ground-truth labels with the model's predictions. 
This matrix is normalised with respect to the ground truth to estimate the bias between classes.

Subsequently, we employ EMA to iteratively update a global matrix, which represents a more stable and comprehensive approximation of the model's class biases. 
The EMA's utility extends beyond smoothing; it removes the need for each class's presence in every batch, simplifying the training process.
The process for constructing this matrix is outlined in Algorithm \ref{algo:icrm}.
The ICRm is formulated separately for both source images, referencing actual ground-truth labels, and target images, utilising pseudo-labels to mirror the ground-truth.
%

\subsection{Class-Relation Augmentation}
In classification task, oversampling is a common technique to counter class imbalance by increasing the presence of minority images.
However, this approach presents challenges in object detection, where images often contain a mix of multiple object classes.
Our analysis on the Cityscapes \cite{cityscapes} dataset indicates that most images include at least one instance of a majority class, rendering image-level resampling ineffective.
This complexity demands more nuanced augmentation strategies for object detection.

Similar to Zhang et al. \cite{Zhang_Pan_Wang_2022}, we employ instance-level oversampling. 
Instances are cropped from their images using bounding box annotations and are then strategically inserted into other images. 

Images are randomly selected from each batch and utilising the ICRm, we differentiate classes as majority or minority based on their likelihood of correct classification.
We then derive the mean probability :
\begin{equation}
{\rm ICRm}_{\rm avg} = \frac{1}{C}\sum^C_{c=0}{\rm ICRm}(c,c),
\end{equation}
where $C$ is the number of classes.
Classes with a probability, ${\rm ICRm}(c,c)$, above and below the mean probability are designated as majority and minority classes, respectively.

Instead of random overlaying, which has been adopted by other previous approaches, our method matches highly related minority and majority instances and uses MixUp \cite{mixup} to blend them, allowing the model to have better generalisation towards minority classes.

We achieve this by pairing each base instance in an image with a sampled instance chosen through weighted random sampling, using the ICRm class probabilities as weights.
For majority base instances, the corresponding column in ICRm, namely,  ${\rm ICRm}(:,c)$ is used, discounting the class's own probability by setting ${\rm ICRm}(c,c)$ to zero to avoid self-augmentation.
This allows us to select classes that are frequently misclassified as the majority class.
%
Conversely, for minority base instances, we use the corresponding row from ICRm as weights without zeroing out ${\rm ICRm}(c,c)$, allowing for the possibility of self-augmentation, which can be beneficial for minority classes.
This is demonstrated in Figure \ref{fig:arch}.b.

Sampled instances are resized to match the base instance dimensions for bounding box consistency. 
The MixUp augmentation is then applied as per the following formulation, where a beta distribution determines the mixing ratio:
\begin{equation}
\begin{aligned}
\hat{I} &= \beta \cdot I_{\rm base} + (1-\beta) \cdot I_{\rm mix},\\
\hat{c} &= \beta \cdot c_{\rm base} + (1-\beta) \cdot c_{\rm mix},
\end{aligned}
\end{equation}
where $I_{\rm base}$ and $I_{\rm mix}$ represent the cropped images of the base and mixed instances, respectively, with $\hat{I}$ denoting the resulting augmented image. 
Similarly, $c_{\rm base}$ and $c_{\rm mix}$ refer to the classes of the base and mixed instances, while $\hat{c}$ indicates the class vector of the augmented instance.

For source domain images, we incorporate instances from both domains to leverage accurate source labels and to aid domain adaptation with target domain samples. 
Whereas for target domain images, we prioritise target instances, using source instances only when no target instances are available for a specific class. 
This ensures that more focus is put onto the target domain for stronger domain adaptation.
Additionally, we do not apply augmentations to minority base instances in the target domain to preserve their integrity.
This ensures that the model is able to focus on the target domain and does not drift to an intermediate domain.

\begin{algorithm}[t]
\caption{Inter-Class Relation module (ICRm)}\label{algo:icrm}
\begin{algorithmic}

\REQUIRE Global class-relation matrix $ICRm$ with shape $(C,C)$, where each element equals $0$.
\WHILE {$training$}
\STATE Create batch matrix ${\rm ICRm}_l$ with shape $(C,C)$, where each element equals $0$.
\FOR {ground-truth, $c_i$, and prediction,$x_i$, in $Y,X$}
	\STATE ${\rm ICRm}_l[c_i,x_i] \gets {\rm ICRm}_l[c_i,x_i] + 1$
\ENDFOR
\FOR {each class $c$ in $C$}
	\STATE Normalise class matrix ${\rm ICRm}_l[c]$ 
	\IF {global class matrix ${\rm ICRm}[c]$ is empty}
		\STATE //Copy batch class matrix to global 
		\STATE ${\rm ICRm} \gets {\rm ICRm}_l$
	\ELSE
		\STATE  //Update global class matrix with EMA
		\STATE  ${\rm ICRm} \gets \beta*{\rm ICRm} + 1(1-\beta)*{\rm ICRm}_l$
	\ENDIF
\ENDFOR
\ENDWHILE
\end{algorithmic}
\end{algorithm}

To implement Class-Relation Augmentation, we store class-specific instance crops from each batch, which we term a Cropbank \cite{Zhang_Pan_Wang_2022}. 
These crops are extracted from bounding box annotations of labelled source images and pseudo-labelled target images.
Separate Cropbanks are maintained for both the source and target datasets, allowing for more targeted augmentation. 
To ensure a diverse range of samples, we update the class instances on a first-in-first-out basis. 
This is particularly beneficial for the target Cropbank, where earlier samples may less accurate due to the early pseudo-labels' robustness.

\subsection{Inter-Class Loss}
To further mitigate class bias, we introduce a weighted parameter to the classification loss, informed by the Inter-Class Relation module (ICRm) for foreground classes. 
This weighting prioritises classes that are frequently misclassified as majority classes, allowing the model to concentrate on refining their performance.
To emphasise the focus on underperforming classes, we employ a non-linear transformation on the ICRm values:
\begin{equation}
    w_i= 
\begin{cases}
    \sqrt{1-{\rm ICRm}(c_i,x_i)} ,& \text{if } c_i = x_i \\
    \sqrt{{\rm ICRm}(c_i,x_i) / {\rm ICRm}(c_i,c_i)},              & \text{otherwise,}
\end{cases}
\label{icl}
\end{equation}
where $w_i$ is the $i$th weight in $W$, and $c_i$ and $x_i$ are the $i$th ground-truth and predicted class, respectively. 
We normalise on the diagonal when $c_i \neq x_i$ as our primary objective is to prioritise low performing classes.
Weights for background classes are uniformly set to 1 to avoid biasing the model against them.
To reconcile the disparity between foreground and background class weights, the weights of the foreground instances are normalised so that their mean equals the background class weight:
\begin{equation}
W_{f} = W_f/{\rm mean}(W_f),
\end{equation}
where $W_f$ denotes the collection of foreground instance weights. 
Moreover, we integrate an additional regularisation term, $\lambda_l$,  across all class-relation weights to prevent extreme weight values from distorting the loss:
\begin{equation}
W = \frac{(W + \lambda_l)}{(1 + \lambda_l)}
\label{eqn_reg}
\end{equation}
This regularisation ensures a moderated, balanced impact on the classification loss, which is now defined as:
\begin{equation}
\mathcal{L}_{\rm cls} = \frac{1}{N}\sum^N_{i=0} w_i * {\rm CE}(c_i,x_i)
\end{equation}
where $N$ is the number of instances and $\rm CE$ is the cross-entropy loss.

The overall loss is then:
\begin{equation}
\mathcal{L} = \mathcal{L}_{\rm sup} + \lambda_{u}\mathcal{L}_{\rm unsup} + \lambda_{d}\mathcal{L}_{\rm dis},
\end{equation}
where $\lambda_{u}$ and $\lambda_{d}$ represent the weights for the unsupervised and discriminator losses, respectively.
\section{Experiments}
\label{sec:experiments}

\begin{table*}[t]
\setlength\tabcolsep{3pt}
\renewcommand{\arraystretch}{1.3}
\fontsize{10pt}{10pt} \selectfont
\centering
\begin{tabularx}{\linewidth}{@{} c|  C{5em} | *{8}{L} |C{5em} @{}} 
\topline
Method & Detector & person & rider & car & truck & bus & train & mcycle & bicycle & mAP \\
\midline
Source \cite{fcos} & FCOS & 36.9 & 36.3 & 44.1 & 18.6 & 29.3 & 8.4 & 20.3 & 31.9 & 28.2 \\
SIGMA \cite{sigma} & FCOS &  46.9 & 48.4 & 63.7 & 27.1 & 50.7 & 35.9 & 34.7 & 41.4 & 43.5 \\
OADA \cite{oada} & FCOS & 47.8 & 46.5 & 62.9 & 32.1 & 48.5 & 50.9 & 34.3 & 39.8 & 45.4 \\
HT \cite{ht} & FCOS & 52.1 & 55.8 & 67.5 & 32.7 & 55.9 & 49.1 & 40.1 & 50.3 & 50.4 \\
\midline
Source \cite{defdetr} & Def DETR & 37.7 & 39.1 & 44.2 & 17.2 & 26.8 & 5.8 & 21.6 & 35.5 & 28.5 \\
AQT \cite{aqt} & Def DETR & 49.3 & 52.3 & 64.4 & 27.7 & 53.7 & 46.5 & 36.0 & 46.4 & 47.1 \\
MRT \cite{mrt} & Def DETR & \textbf{52.8} & 51.7 & \textbf{68.7} & 35.9 & 58.1 & \textbf{54.5} & 41.0 & 47.1 & 51.2 \\
\midline
Source \cite{frcnn} & FRCNN & 22.4 & 26.6 & 28.5 & 9.0 & 16.0 & 4.3 & 15.2 & 25.3 & 18.4 \\
Oracle & FRCNN & 39.5 & 47.3 & 59.1 & 33.1 & 47.3 & 42.9 & 38.1 & 40.8 & 43.5 \\
MeGA \cite{mega} & FRCNN &  37.7 & 49.0 & 52.4 & 25.4 & 49.2 & 46.9 & 34.5 & 39.0 & 41.8 \\
TIA \cite{tia}  & FRCNN &  34.8 & 46.3 & 49.7 & 31.1 & 52.1 & 48.6 & 37.7 & 38.1 & 42.3 \\
UMT \cite{umt} & FRCNN  & 33.0 & 46.7 & 48.6 & 34.1 & 56.5 & 46.8 & 30.4 & 37.3 & 41.7 \\
TDD \cite{tdd} & FRCNN  & 39.6 & 47.5 & 55.7 & 33.8 & 47.6 & 42.1 & 37.0 & 41.4 & 43.1 \\
PT \cite{pt}  & FRCNN &  40.2 & 48.8 & 59.7 & 30.7 & 51.8 & 30.6 & 35.4 & 44.5 & 42.7 \\
AT‡  \cite{at} & FRCNN & 45.3 & 55.7 & 63.6 & 36.8 & 64.9 & 34.9 & 42.1 & 51.3 & 49.3 \\
CMT  \cite{cmt} & FRCNN &  45.9 & 55.7 & 63.7 & 39.6 & \textbf{66.0} & 38.8 & 41.4 & 51.2 & 50.3\\
MILA \cite{mila} & FRCNN & 45.6 & 52.8 & 64.8 & 34.7 & 61.4 & 54.1 & 39.7 & 51.5 & 50.6 \\
\midline
CAT (Ours) & FRCNN & 44.6 & \textbf{57.1} & 63.7 & \textbf{40.8} & \textbf{66.0} & 49.7 & \textbf{44.9} & \textbf{53.0} &\textbf{52.5} \\
\end{tabularx}
\caption{Object detection results on the Foggy Cityscapes test set for \textbf{Cityscapes $\to$ Foggy Cityscapes (0.02) domain adaptation}. We group methods based on their detector frameworks (FCOS/Def DETR/FRCNN) and highlight the best performing method. CAT is able to outperform the previous state-of-the-art, MRT, by 1.3 mAP and improve on AT by 3.2 mAP. The mean average precision at .50 IoU (mAP) is reported for all classes. ‡ AT performance is reproduced on Foggy Cityscapes (0.02) with publicly available code for fairness.}
\label{table_foggy}
\end{table*}

\subsection{Datasets}

We assess the performance of Class-Aware Teacher (CAT) using benchmarks in domain adaptive object detection (DAOD) following prior work \cite{at}.

\paragraph{Cityscapes $\to$ Foggy Cityscapes:} 
The Cityscapes dataset \cite{cityscapes} is a road-centric dataset with 2,975 training and 500 validation images from various urban settings under clear weather, annotated across 8 classes.
Foggy Cityscapes \cite{foggycityscapes} is a synthesised dataset generated on Cityscapes to simulate foggy weather, using the same base images and annotations.  
We conduct our experiment on the most severe fog level (0.02) where Foggy Cityscapes is used as the target domain.

\paragraph{PASCAL VOC $\to$ Clipart1K:} 
We use PASCAL VOC 2012 \cite{pascal}, which comprises of 11,540 real-world images across 20 categories, for training. 
The Clipart1k dataset \cite{clipart} consists of 20 corresponding clipart object categories. 
Following \cite{at}, we split Clipart1k into 500 training and 500 testing images.

\vspace{0.3cm}
\noindent The benchmarks of Sim10K \cite{sim10k} $\to$ Cityscapes and KITTI \cite{kitti} $\to$ Cityscapes are excluded from our evaluation. 
Despite their popularity in DAOD research, they focus solely on the 'Car' category, which does not align with our class imbalance setting.

\subsection{Experimental Setup}

Following previous works in DAOD, we adopt the Faster R-CNN object detector with VGG-16 \cite{vgg} and ResNet-101 \cite{resnet} as backbones for our detection model.
Our hyperparameters: EMA decay rate $\alpha = 0.9996$, beta-distribution hyper parameters [0.5,0.5], adversarial loss weight, $\lambda_{d}$ $0.1$, and unsupervised loss weight, $\lambda_{u}$ $1.0$, regularisation term, $\lambda_{l}$ 1.0. 
We employ a hard threshold $\tau$ of 0.8 for pseudo-labelling.
Weak-strong augmentation \cite{unbiased} is applied to both source and target images. 
We train our student model for 20,000 iterations on the labelled source data.
The parameters of the student is copied to the teacher which is then updated via EMA of the student at each iteration.
We continue training for 60,000 iterations with both labelled source and unlabelled target data.
Our framework is developed on top of the publicly available Detectron2 \cite{detectron2}. 
Experiments are performed using a batch size of 8 source and 8 target images, distributed across 4 NVIDIA RTX3090 GPUs.
Additional details regarding our experimental setup are provided in the supplementary materials.

\begin{table*}[t]
\setlength\tabcolsep{3pt}
\renewcommand{\arraystretch}{1.3}
\fontsize{9pt}{9pt} \selectfont
\centering
\begin{tabularx}{\linewidth}{@{} c| *{20}{L} |C{3em} @{}} \hline
Method &  aero & bike & bird & boat & bottle & bus & car & cat & chair & cow & table & dog & horse & mtr & prsn & plant & shp & sofa & train & tv & mAP \\ \hline
Source \cite{fastrcnn} & 23.0 & 39.6 & 20.1 & 23.6 & 25.7 & 42.6 & 25.2 & 0.9 & 41.2 & 25.6 & 23.7 & 11.2 & 28.2 & 49.5 & 45.2 & 46.9 & 9.1 & 22.3 & 38.9 & 31.5 & 28.8 \\
Oracle & 33.3 & 47.6 & 43.1 & 38.0 & 24.5 & 82.0 & 57.4 & 22.9 & 48.4 & 49.2 & 37.9 & 46.4 & 41.1 & 54.0 & 73.7 & 39.5 & 36.7 & 19.1 & 53.2 & 52.9 & 45.0 \\
\midline
I3Net \cite {i3net} & 30.0 & \textbf{67.0} & 32.5 & 21.8 & 29.2 & 62.5 & 41.3 & 11.6 & 37.1& 39.4& 27.4& 19.3& 25.0& 67.4& 55.2& 42.9& 19.5& 36.2& 50.7& 39.3 &37.8 \\
ICR-CCR \cite{Xu2020ExploringCR} & 28.7 & 55.3 & 31.8 & 26.0 & 40.1 & \textbf{63.6} & 36.6 & 9.4 & 38.7 & 49.3 & 17.6 & 14.1 & 33.3 & 74.3 & 61.3 & 46.3 & 22.3 & 24.3 & 49.1 & 44.3 & 38.3 \\
HTCN \cite{htcn} & 33.6 & 58.9 & 34.0 & 23.4 & 45.6 & 57.0 & 39.8 & 12.0 & 39.7 & 51.3 & 20.1 & 20.1 & 39.1 & 72.8 & 61.3 & 43.1 & 19.3 & 30.1 & 50.2 & 51.8 & 40.3 \\
DM \cite{dm} & 25.8 & 63.2 & 24.5 & 42.4 & 47.9 & 43.1 & 37.5 & 9.1 & 47.0 & 46.7 & 26.8 & 24.9 & 48.1 & 78.7 & 63.0 & 45.0 & 21.3 & 36.1 & 52.3 & \textbf{53.4} & 41.8 \\
UMT \cite{umt}& 39.6 & 59.1 & 32.4 & 35.0 & 45.1 & 61.9 & 48.4 & 7.5 & 46.0 & \textbf{67.6} & 21.4 & \textbf{29.5} & 48.2 & 75.9 & 70.5 & 56.7 & 25.9 & 28.9 & 39.4 & 43.6 & 44.1 \\
TIA \cite{tia}& 42.2 & 66.0 & 36.9 & 37.3 & 43.7 & 71.8 & 49.7 & 18.2 & 44.9 & 58.9 & \textbf{18.2} & 29.1 & 40.7 & 87.8 & 67.4 & 49.7 & \textbf{27.4} & 27.8 & \textbf{57.1} & 50.6 & 46.3 \\
AT‡ \cite{at} & 33.1 & 66.1 & 35.3 & 44.9 & 57.5 & \textbf{44.9} & 51.0 & 5.8 & 59.5 & 54.9 & 34.6 & 23.5 & 64.3 & 84.0 & 75.4 & 51.5 & 17.1 & 30.3 & 43.3 & 37.2 & 45.7 \\
CMT \cite{cmt} & 39.8 & 56.3 & 38.7 & 39.7 & 60.4 & 35.0 & \textbf{56.0} & 7.1 & \textbf{60.1} & 60.4 & 35.8 & 28.1 & \textbf{67.8} & 84.5 & \textbf{80.1} & 55.5 & 20.3 & 32.8 & 42.3 & 38.2 & 47.0 \\
\midline
CAT (Ours) & \textbf{40.5} & 64.1 & \textbf{38.8} & 41.0 & \textbf{60.7} & 55.5 & 55.6 & 14.3 & 54.7 & 59.6 & \textbf{46.2} & 20.3 & 58.7 & \textbf{92.9} & 62.6 & \textbf{57.5} & 22.4 & \textbf{40.9} & 49.5 & 46.0 & \textbf{49.1}\\
\hline
\end{tabularx}
\caption{Object detection results on the Clipart1k test set for \textbf{PASCAL VOC $\to$ Clipart1k domain adaptation}. CAT improves on the previous state-of-the-art, CMT, by 2.1 mAP, achieving the new best of 49.1 mAP. The mean average precision at .50 IoU (mAP) is reported for all classes.‡ AT performance is reproduced following \cite{cmt}.}
\label{pascal_table}
\end{table*}

\subsection{Comparison with SOTA methods}
We compare our method with the state-of-the-art in DAOD as well as reporting a source only FCOS/Def DETR/Faster RCNN for a baseline comparison. 
We additionally include an oracle upper bound, which is trained on only the target domain and its ground truth annotations. 

\paragraph{Foggy Weather Adaptation}
When object detectors are deployed in real-world scenarios, the performance could drop significantly under sub-optimal conditions, e.g. adverse weather.
This is because that the samples in adverse weather do not present in the training of the model resulting in a domain shift.    
The domain adaptation task is designed to overcome this gap between normal and adverse conditions.
To demonstrate this, we conduct an experiment on the commonly used Cityscapes $\to$ Foggy Cityscapes benchmark.

Our results are shown in Table \ref{table_foggy}. We can observe that methods utilising the student-teacher framework (HT, UMT, TDD, PT, AT, CMT, MILA, MRT) outperform non-student-teacher frameworks by a large margin. 
CAT, which is built on existing SOTA mean teacher frameworks, significantly improves the performance at 52.5 mAP. 
Additionally, our method is able to improve minority classes while not impacting majority classes.

\paragraph{Real to Artistic Adaptation} 
Adapting object detection from real to artistic domains is particularly challenging due to the significant differences between these domains. 
In our experiment, detailed in Table \ref{pascal_table}, we observe CAT achieves a mAP of 49.1, outperforming the previous best, by 2.1 mAP and AT by 3.4 mAP.

Notably, CAT shows substantial improvements in minority classes, such as 'motorbike' which in the Clipart1k training set consists of only 7 images.
The results of this experiment demonstrate CAT's effectiveness in addressing class imbalances even in dissimilar domains.

%
%
%

\begin{table}[t]
\setlength\tabcolsep{3pt}
\renewcommand{\arraystretch}{1.2}
\fontsize{10pt}{10pt} \selectfont
\centering
\begin{tabularx}{0.9\linewidth}{@{} c|  *{3}{L} | *{2}{L} @{}} 
 \hline
 Method & ICRm & CRA & ICL & mAP & $\sigma \downarrow$ \\
 \hline
Base (AT \cite{at}) &  &  &  &  49.3 & 10.8 \\
\midline
  & \cmark & \cmark &  & 51.0 & 8.8 \\
CAT (Ours) & \cmark & & \cmark & 51.6 & 10.3 \\
 & \cmark & \cmark & \cmark & 52.5 & 8.6 \\
\end{tabularx}
\caption{\textbf{Ablation studies on CAT components.} ICRm is included in all studies as it forms the basis of CRA and ICL. We report the mean average precision at .50 IoU (mAP) and the standard deviation of class-mAP ($\sigma$). Our contributions are not included in the base framework (AT).}
\label{table:abla}
\end{table}

\subsection{Ablation Studies}
To verify the significance of our contributions, we conducted an ablation study. 
All experiments within this study were performed on the Cityscapes $\to$ Foggy Cityscapes benchmark using the VGG16 backbone.
\paragraph{Quantitative Ablation}
Table \ref{table:abla} quantitatively showcases the efficacy of each contribution within our framework. 
The base framework, prior to integrating our modules, corresponds to the AT model as described in \cite{at}. 
Since our Inter-Class Relation module (ICRm) is pivotal for both the class-relation augmentation and loss, it is a constant across all experimental variations.
To highlight our method's capability in addressing class imbalance, we introduce $\sigma$. 
This value represents the standard deviation of the mAP across different classes and serves as an indicator of performance equity among classes.

Inclusion of our Inter-Class Loss (ICL) yields a notable 2.3 mAP gain over the base AT model and a decreased $\sigma$, indicating a more balanced performance across classes.
The Class Relation Augmentation (CRA) also benefits AT, though to a lesser extent than ICL in terms of mAP. 
Notably, CRA significantly narrows the performance gap between minority and majority classes, as reflected by a reduced $\sigma$ of 8.8.
Employing both ICL and CRA not only enhances overall performance but also achieves a lower $\sigma$ compared to the base model, reinforcing our method's effectiveness in managing class imbalance.

\begin{figure*}[t]
\centering
\includegraphics[width=0.95\textwidth]{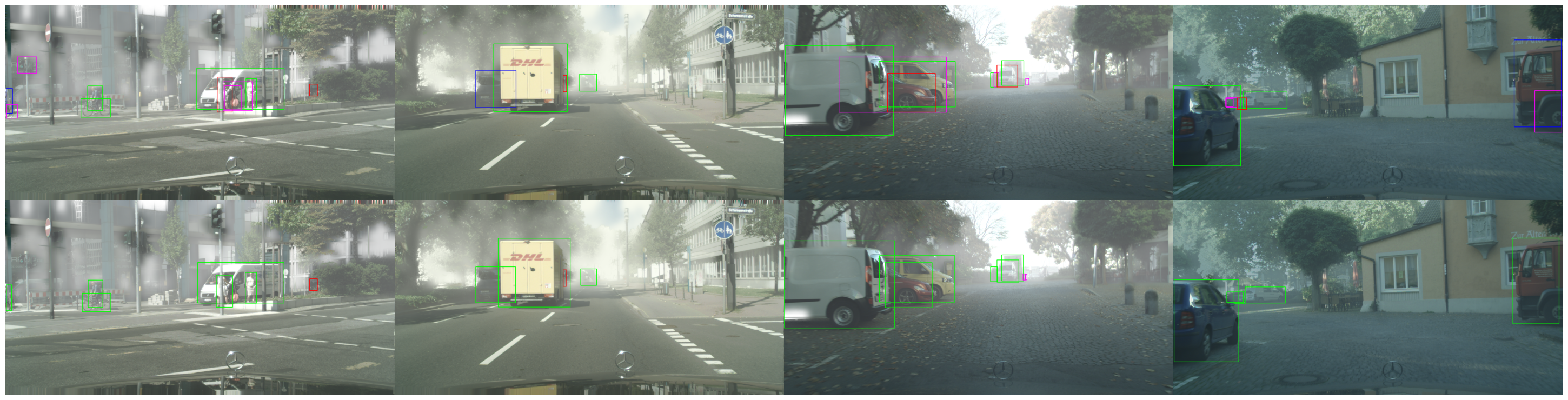}
\caption{\textbf{Qualitative results of CAT.} We show the results of AT and CAT on the top and bottom, respectively. CAT is able to address misclassification (col 1,2,4), false negatives (col 1,3), and false positives (col 1,3,4). Box colour represents: Green $\to$ true positives, Blue $\to$ misclassified, Red $\to$ false negatives, Pink $\to$ false positives.}
\centering
\label{qual}
\end{figure*}

\begin{table}[t]
\setlength\tabcolsep{3pt}
\renewcommand{\arraystretch}{1.1}
\fontsize{10pt}{10pt} \selectfont
\centering
\begin{tabularx}{0.7\linewidth}{@{} c| *{2}{L} @{}} 
 \hline
Selection Method & mAP & $\sigma \downarrow$\\
\midline
Random (0.5) & 51.8 & 8.5\\
CRA (1.0) & 50.2 & 9.2 \\
CRA (0.5) & 52.5 & 8.6\\
\end{tabularx}
\caption{Comparison of selecting class instances randomly and via CRA. Values in brackets refer to the likelihood of an instance being augmented.}
\label{table:mixup_abla}
\end{table}

\paragraph{Impact of Augmentation}
We demonstrate the impact of augmentation in terms of ratio and selection criteria.
Augmenting images is a key part of our approach, enriching the dataset with additional representations of minority classes.
Yet, if images are augmented excessively, the model may fail to learn an accurate representation of the classes.
To strike a balance, we selectively augment a random subset of the images.
Additionally, how class instances are paired is key to improve the quality of augmentation.
Compared to randomly selecting class pairs, CRA is able to prioritise pairing highly related minority and majority class instances.
This ensures that the Mixup output is more meaningful for minority class performance.

The experimental results of this approach are given in Table \ref{table:mixup_abla}.
We compare randomly selecting class instances for MixUp with our Class-Relation Augmentation at different values.
These values represent the likelihood of a instance in the base image being augmented.
Randomly applying MixUp improves the overall performance by +0.2 mAP.
However, by using CRA we can further increase performance by +0.9 mAP.
This shows that pairing highly-related classes for MixUp strengthens the performance of the minority class while minimally affecting the majority class.
In addition, our experiments show that too much augmentation can have a negative effect on the performance of the model.
%

\paragraph{Weighing Strategy for Class Loss}
We introduce a weighted classification loss to improve the performance of minority class performance in Eq. \ref{icl}. 
Class-level loss is a common strategy that has been adopted to address the class imbalance in a dataset \cite{Huang_2016_CVPR,NIPS2017_147ebe63}.
We compare our Inter-Class Loss (ICL) with a variant of previous class-level losses where only the diagonal of the Inter-Class Relation module is used.
The diagonal corresponds to the ground-truth class likelihood of accurate classification.
In contrast, ICL uses the likelihood of the ground-truth being classified as the predicted class to influence its loss.
We show in Table \ref{table:weight_alba} that by using this inter-class relationship, we are able to improve the performance by +0.5 mAP.
However, there may be cases where ICL overly penalises well-performing classes.
This is addressed by applying a regularisation term as seen in Eq. \ref{eqn_reg}.
We can see that if this regularisation term is removed, performance drops significantly as the weight of certain classes gets too small during training.

\begin{table}[t]
\setlength\tabcolsep{3pt}
\renewcommand{\arraystretch}{1.1}
\fontsize{10pt}{10pt} \selectfont
\centering
\begin{tabularx}{0.7\linewidth}{@{} c| *{2}{L} @{}} 
 \hline
Class Weight & mAP & $\sigma \downarrow$\\
\midline
Class-Level & 52.0 & 9.0\\
ICL w/o Reg. & 51.3 & 9.5\\
ICL & 52.5 & 8.6\\
\end{tabularx}
\caption{Class Loss Weighing Strategies. Class-level only uses the diagonal values in our ICRm along with regularisation, ICL refers to our Inter-Class Loss. }
\label{table:weight_alba}
\end{table}

\paragraph{Qualitative Results}
Figure \ref{qual} illustrates the qualitative results of our method, with the top and bottom row displaying predictions from AT and CAT, respectively.

CAT is able to correct misclassifications, represented by blue boxes.
Additionally, CAT can bring a reduction in both false positives and false negatives, represented by pink and red boxes, showcasing improved detection accuracy across various scales and classes.
%
%
%
%

\section{Conclusion}
\label{sec:conclusion}
In this paper, we propose Class-Aware Teacher (CAT) for Domain Adaptive Object Detection.
We demonstrate that CAT, by leveraging our Inter-Class Relation module, effectively approximates and mitigates class biases, leading to more equitable performance across classes.
Furthermore, Class-Relation Augmentation and Inter-Class Loss were shown to be effective in enhancing minority class representation.
Our experimental results on Cityscapes → Foggy Cityscapes and PASCAL VOC → Clipart1K have demonstrated the effectiveness of our method achieving SOTA performance at 52.5 mAP and 49.1 mAP, respectively.
Based on our findings, we believe that further investigation into inter-class dynamics is a promising direction for advancing class imbalance in the DAOD setting.

{
    \small
    \bibliographystyle{ieeenat_fullname}
    \bibliography{main}
}

\input{sec/X_suppl.tex}

\end{document}

%% file: preamble.tex
\usepackage[dvipsnames]{xcolor}


%% file: sec/X_suppl.tex
\clearpage
\setcounter{page}{1}
\maketitlesupplementary

\section{Additional Details on Methods}

\subsection{Additional Details on Class-Relation Augmentation}
We further describe the details of our Class-Relation Augmentation (CRA) approach below.
CRA augments random images in a batch based on the source and target augmentation ratio.
For each selected image, we identify class instances using labels or pseudo-labels for source and target images, respectively, termed 'base instances.'

Following the methodology outlined in Section 4.2, we select 'mix instances' that exhibit a strong relationship with the base instances, determined by our Inter-Class Relation module (ICRm). 
A 'mixed instance' is then randomly chosen from a predefined crop bank. 
To mitigate the effects of upsampling degradation, we ensure the mixed instances is at least 0.25 of the base instance's size.

We resize the mixed instance to the base instance's dimensions, allowing the aspect ratio of the mixed instance to be adjusted.
This resizing allows us to use a single bounding box to represent both the base and mixed instance after augmentation.
Experimental results, presented in Table 1, demonstrate that this resizing strategy not only maintains but enhances model performance compared to maintaining the mixed instance aspect ratio.
This is because the ambiguity of labelling when two bounding boxes are used is complex, especially when employing mixup.
Once the mixed instance has been resized, mixup \cite{mixup} is then applied to combine the two instances and their labels.
Given the distinct class representations, we employ one-hot encoding to support multi-class labelling. 
This process is repeated across all objects in the selected image.

\begin{table}[H]
\setlength\tabcolsep{3pt}
\renewcommand{\arraystretch}{1.1}
\fontsize{10pt}{10pt} \selectfont
\centering
\begin{tabularx}{0.7\linewidth}{@{} L| *{1}{L} @{}} 
 \hline
Maintain Aspect Ratio & mAP \\
\midline
\xmark & 52.5 \\
\cmark & 51.1 \\
\end{tabularx}
\caption{Performance of Class-Aware Teacher (CAT) with and without maintaining the aspect ratio during CRA. We can see that disregarding the aspect ratio during resizing improves performance while being a simpler resizing strategy.}
\label{table:weightstrat}
\end{table}

\section{Experiments}
\subsection{Additional Details on Experimental Setup}
\label{sec:rationale}
In this section, we provide additional details on the experimental setup. 
Consistent with prior research in the domain of adaptive object detection, our experiments are conducted using the Faster R-CNN detection framework.
VGG-16 \cite{vgg} and ResNet-101 \cite{resnet} are used as the backbones for our detection model depending on the benchmark used.
PASCAL VOC $\to$ Cliapart1K utilises the ResNet-101 backbone.
Both Cityscapes $\to$ Foggy Cityscapes and Cityscapes $\to$ BDD100K utilises the VGG-16 backbone.

Across all experiments, we maintain consistent hyperparameter settings, which are detailed in Table 2. 

\subsection{Additional Details on Dataset Class Distributions}

The distribution of classes in our datasets plays an important role during training. 
Minority classes tend to under perform, especially when there is a distribution shift between training and validation datasets. 
To validate the effectiveness of our method, we show the class distributions of the evaluation datasets and how our method is able to address minority class performance.

\begin{figure}[t]
\includegraphics[width=0.5\textwidth]{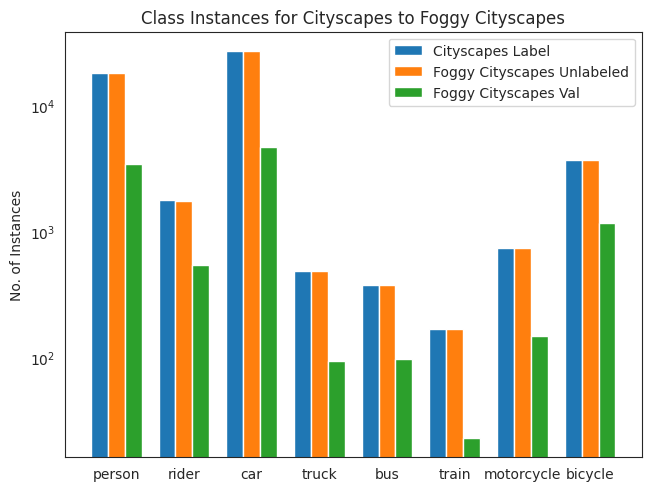}
\caption{Class distribution of datasets used for the Cityscapes $\to$ Foggy Cityscapes task. We can see that person and car classes form the majority of all classes. The distribution of classes for the labeled dataset and validation set is similar which makes for an simpler task.}
\centering
\label{fig:cs2fcs}
\end{figure}

Figure \ref{fig:cs2fcs} shows the class distribution for the Cityscapes $\to$ Foggy Cityscapes task. 
Car and person forms the majority in all the datasets used for this task and truck, bus, and train form the minority. 
This is to be expected as the datasets are from the same source and would share similar distributions.
This forms a simpler task as we do not need to account for a distribution shift during testing.
Our method matches or outperforms SOTA for the truck and bus class, as well as strongly outperforming our base method \cite{at} for all three minority classes.

\begin{figure}[t]
\includegraphics[width=0.5\textwidth]{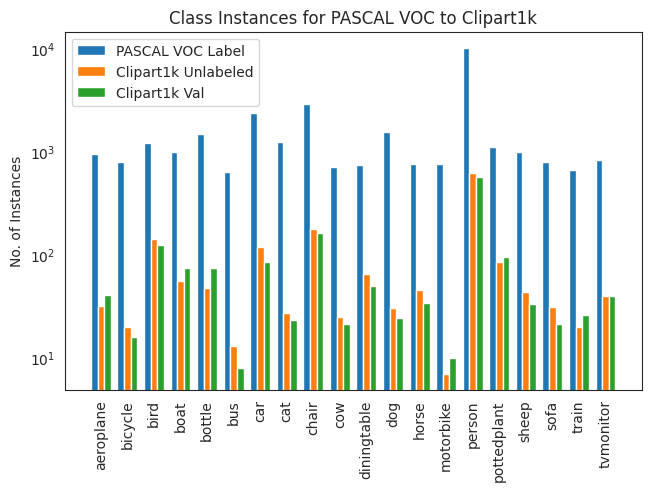}
\caption{Class distribution of datasets used for the PASCAL VOC $\to$ Clipart1k task. Person is a majority class for all datasets, however other classes for PASCAL VOC have a similar number of instance. The imbalance is stronger in the Clipart1K dataset with classes such as motorbike and bus being a minority.}
\centering
\label{fig:voc2clip}
\end{figure}

The class distribution of the PASCAL VOC $\to$ Clipart1k task is shown in Figure \ref{fig:voc2clip}.
The PASCAL VOC dataset is fairly balanced with the only outlier being the person class.
This ensures that the initial training has less bias towards specific classes, however, Clipart1k exhibits stronger class imbalance.
This results in a distribution shift during unsupervised training and evaluation which may result in suboptimal performance.
CAT is able to have strong performance on the motorbike minority class and is able to outperform its base on the bus class.

\begin{figure}[t]
\includegraphics[width=0.5\textwidth]{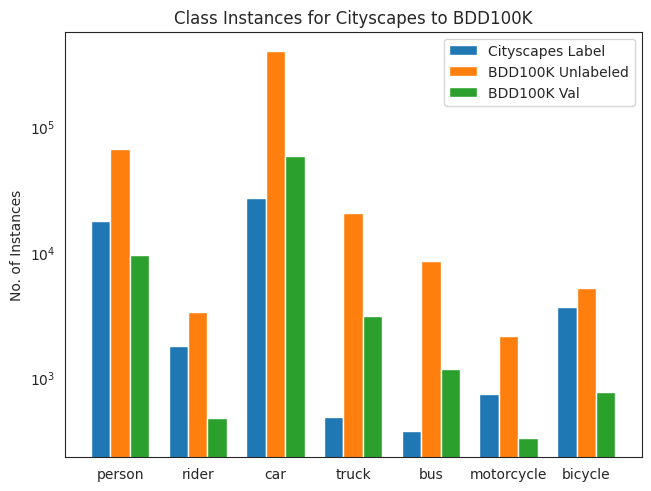}
\caption{Class distribution of datasets used for the Cityscapes $\to$ BDD100K (Daytime) task. We can see that the car class form the majority of all classes, especially for the BDD100K dataset. Note that the class distribution of labeled and validation set differs, especially for minority classes which can make the task more difficult.}
\centering
\label{fig:cs2bdd}
\end{figure}

The Cityscapes $\to$ BDD100K (Daytime) task contains two road-centric datasets taken in different locations which would result in both imbalanced data as well as a distribution shift as seen in Figure \ref{fig:cs2bdd} .
This would be a harder task as a minority class in one dataset may not be the same minority the other. 
For example, truck and bus are the minority for Cityscapes but motorcycle and bicycle are the minority for BDD100K. 
CAT is able to outperform SOTA for truck, bus, and bicycle and is only 0.1 mAP lower for the motorcycle minority class.

\subsection{Cityscapes $\to$ BDD100K}
In addition to experiments performed in Section 5.3 of the main paper, we include the Cityscapes $\to$ BDD100K-Daytime benchmark. 

The BDD100K \cite{bdd100k} dataset is a large-scale dataset containing 100,000 images. For this experiment, we use the day-time split which contains 36,728 training and 5,258 testing images. We remove the train, traffic light and traffic sign categories following previous work. The Cityscapes $\to$ BDD100K benchmark covers scene adaptation as well as small-to-large dataset adaptation.

Table \ref{table:cs2bdd} shows the results of our experiment. 
We can observe that CAT has stronger performance compared to the previous SOTA at 38.5 mAP.
Minority classes such as rider, truck, bus, and bicycle also show a significant improvement.
This shows that our strategy to address inter-class dynamics provides a viable solution to address class imbalance for domain adaptive object detection.

\begin{table*}[t]
\centering
\begin{tabularx}{0.96\linewidth}{@{} c| *{7}{L} |C{5em} @{}} 
\hline
Method & person & rider & car & truck & bus & mcycle & bicycle & mAP \\ 
\hline
Faster RCNN \cite{frcnn} & 28.8 & 25.4 & 44.1 & 17.9 & 16.1 & 13.9 & 22.4 & 24.1 \\
SIGMA \cite{sigma} & 46.9 & 29.6 & 64.1 & 20.2 & 23.6 & 17.9 & 26.3 & 32.7 \\
TDD \cite{tdd}  & 39.6 & 38.9 & 53.9 & 24.1 & 25.5 & 24.5 & 28.8 & 33.6 \\
PT \cite{pt}  & 40.5 & 39.9 & 52.7 & 25.8 & 33.8 & 23.0 & 28.8 & 34.9 \\
\midline
CAT (Ours) & 44.6 & \textbf{41.5} & 61.2 & \textbf{31.4} & \textbf{34.6} & 24.4 & \textbf{31.7} & \textbf{38.5} \\
\end{tabularx}
\caption{Object detection results on the BDD100k-Daytime test set for \textbf{Cityscapes $\to$ BDD100k-Daytime domain adaptation}. The mean average precision at .50 IoU (mAP) is reported for all classes.}
\label{table:cs2bdd}
\end{table*}

\begin{table*}[t]
\setlength\tabcolsep{3pt}
\renewcommand{\arraystretch}{1.3}
\fontsize{10pt}{10pt} \selectfont
\centering
\begin{tabularx}{0.90\linewidth}{@{} *{1}{L} c *{3}{L}  @{}} 
 \hline
 Hyperparameter & Description & C$\to$F & PV$\to$CA & C$\to$B \\
\midline
- & Detector & FRCNN & FRCNN & FRCNN \\
- & Backbone & VGG & ResNet-101 & VGG \\
- & BatchNorm & True & True & False \\
$\alpha$ & Decay rate for student-teacher EMA & 0.9996 & 0.9996 & 0.9996 \\
$\beta$ & Beta-distribution parameters for mixup & [0.5,0.5] & [0.5,0.5] & [0.5,0.5] \\
$\lambda_d$ & Weight for Adverserial Loss & 0.1 & 0.1 & 0.1 \\
$\lambda_u$ & Weight for Unsupervised Loss & 1.0 & 1.0 & 1.0 \\
$\tau$ & Threshold value for pseudo-label confidence & 0.8 & 0.8 & 0.8 \\
$\lambda_l$ & Regularization term for Inter-Class Loss & 1.0 & 1.0 & 1.0 \\
- & Source Augmentation Ratio & 0.5 & 0.5 & 0.5 \\
- & Target Augmentation Ratio & 0.5 & 0.5 & 0.5 \\
- & Burn-Up Step Iterations & 20000 & 20000 & 20000 \\
- & Total Training Iterations & 80000 & 80000 & 80000 \\
- & Learning Rate & 0.2 & 0.2 & 0.2 \\

\end{tabularx}
\caption{Model Hyperparameters for Experiments.From left to right, Cityscapes $\to$ Foggy Cityscapes, PASCAL VOC $\to$ Clipart1K, and Cityscapes $\to$ BDD100K-Day.}
\label{table:setup}
\end{table*}

%% file: v1_mikhail.bbl
\begin{thebibliography}{52}
\providecommand{\natexlab}[1]{#1}
\providecommand{\url}[1]{\texttt{#1}}
\expandafter\ifx\csname urlstyle\endcsname\relax
  \providecommand{\doi}[1]{doi: #1}\else
  \providecommand{\doi}{doi: \begingroup \urlstyle{rm}\Url}\fi

\bibitem[Cao et~al.(2023)Cao, Joshi, Gui, and Wang]{cmt}
Shengcao Cao, Dhiraj Joshi, Liang-Yan Gui, and Yu-Xiong Wang.
\newblock Contrastive mean teacher for domain adaptive object detectors.
\newblock In \emph{IEEE/CVF Conference on Computer Vision and Pattern
  Recognition}, pages 23839--23848, 2023.

\bibitem[Chen et~al.(2020)Chen, Zheng, Ding, Huang, and Dou]{htcn}
Chaoqi Chen, Zebiao Zheng, Xinghao Ding, Yue Huang, and Qi Dou.
\newblock Harmonizing transferability and discriminability for adapting object
  detectors.
\newblock In \emph{IEEE/CVF Conference on Computer Vision and Pattern
  Recognition}, 2020.

\bibitem[Chen et~al.(2021)Chen, Zheng, Huang, Ding, and Yu]{i3net}
Chaoqi Chen, Zebiao Zheng, Yue Huang, Xinghao Ding, and Yizhou Yu.
\newblock I3net: Implicit instance-invariant network for adapting one-stage
  object detectors.
\newblock In \emph{IEEE/CVF Conference on Computer Vision and Pattern
  Recognition}, pages 12576--12585, 2021.

\bibitem[Chen et~al.(2022)Chen, Chen, Yang, Song, Wang, Zhang, Yan, Qi, Zhuang,
  Xie, et~al.]{pt}
Meilin Chen, Weijie Chen, Shicai Yang, Jie Song, Xinchao Wang, Lei Zhang,
  Yunfeng Yan, Donglian Qi, Yueting Zhuang, Di Xie, et~al.
\newblock Learning domain adaptive object detection with probabilistic teacher.
\newblock In \emph{International Conference on Machine Learning}, pages
  3040--3055, 2022.

\bibitem[Chen et~al.(2018)Chen, Li, Sakaridis, Dai, and Gool]{dafaster}
Yuhua Chen, Wen Li, Christos Sakaridis, Dengxin Dai, and Luc~Van Gool.
\newblock Domain adaptive faster r-cnn for object detection in the wild.
\newblock In \emph{IEEE/CVF Conference on Computer Vision and Pattern
  Recognition}, pages 3339--3348, 2018.

\bibitem[Chou et~al.(2020)Chou, Chang, Pan, Wei, and Juan]{remix}
H. Chou, S. Chang, J. Pan, W. Wei, and D. Juan.
\newblock Remix: rebalanced mixup.
\newblock \emph{Computer Vision – ECCV 2020 Workshops}, pages 95--110, 2020.

\bibitem[Cordts et~al.(2016)Cordts, Omran, Ramos, Rehfeld, Enzweiler, Benenson,
  Franke, Roth, and Schiele]{cityscapes}
Marius Cordts, Mohamed Omran, Sebastian Ramos, Timo Rehfeld, Markus Enzweiler,
  Rodrigo Benenson, Uwe Franke, Stefan Roth, and Bernt Schiele.
\newblock The cityscapes dataset for semantic urban scene understanding.
\newblock In \emph{IEEE/CVF Conference on Computer Vision and Pattern
  Recognition}, pages 3213--3223, 2016.

\bibitem[Deng et~al.(2021)Deng, Li, Chen, and Duan]{umt}
Jinhong Deng, Wen Li, Yuhua Chen, and Lixin Duan.
\newblock Unbiased mean teacher for cross-domain object detection.
\newblock In \emph{IEEE/CVF Conference on Computer Vision and Pattern
  Recognition}, pages 4089--4099, 2021.

\bibitem[Deng et~al.(2023)Deng, Xu, Li, and Duan]{ht}
Jinhong Deng, Dongli Xu, Wen Li, and Lixin Duan.
\newblock Harmonious teacher for cross-domain object detection.
\newblock In \emph{IEEE/CVF Conference on Computer Vision and Pattern
  Recognition}, pages 23829--23838, 2023.

\bibitem[Everingham et~al.(2009)Everingham, Gool, Williams, Winn, and
  Zisserman]{pascal}
Mark Everingham, Luc~Van Gool, Christopher K.~I. Williams, John Winn, and
  Andrew Zisserman.
\newblock The pascal visual object classes (voc) challenge.
\newblock \emph{International Journal of Computer Vision}, pages 303--308,
  2009.

\bibitem[Galdrán et~al.(2021)Galdrán, Carneiro, and Ballester]{balmixup}
A. Galdrán, G. Carneiro, and M.~Á.~G. Ballester.
\newblock Balanced-mixup for highly imbalanced medical image classification.
\newblock \emph{Medical Image Computing and Computer Assisted Intervention –
  MICCAI 2021}, pages 323--333, 2021.

\bibitem[Geiger et~al.(2013)Geiger, Lenz, Stiller, and Urtasun]{kitti}
Andreas Geiger, Philip Lenz, Christoph Stiller, and Raquel Urtasun.
\newblock Vision meets robotics: The kitti dataset.
\newblock \emph{International Journal of Robotics Research}, 2013.

\bibitem[Girshick(2015)]{fastrcnn}
Ross Girshick.
\newblock Fast r-cnn.
\newblock In \emph{IEEE/CVF International Conference on Computer Vision}, pages
  1440--1448, 2015.

\bibitem[He and Garcia(2009)]{he2009learning}
Haibo He and Edwardo~A Garcia.
\newblock Learning from imbalanced data.
\newblock \emph{IEEE Transactions on knowledge and data engineering},
  21\penalty0 (9):\penalty0 1263--1284, 2009.

\bibitem[He et~al.(2016)He, Zhang, Ren, and Sun]{resnet}
Kaiming He, Xiangyu Zhang, Shaoqing Ren, and Jian Sun.
\newblock Deep residual learning for image recognition.
\newblock In \emph{IEEE/CVF Conference on Computer Vision and Pattern
  Recognition}, pages 770--778, 2016.

\bibitem[He et~al.(2022)He, Wang, Wu, Wang, Li, Li, Gan, Wu, and Qiao]{tdd}
Mengzhe He, Yali Wang, Jiaxi Wu, Yiru Wang, Hanqing Li, Bo Li, Weihao Gan, Wei
  Wu, and Yu Qiao.
\newblock Cross domain object detection by target-perceived dual branch
  distillation.
\newblock In \emph{IEEE/CVF Conference on Computer Vision and Pattern
  Recognition}, pages 9560--9570, 2022.

\bibitem[Hsu et~al.(2015)Hsu, Chen, Hou, Tsai, Yeh, and
  Wang]{hsuUnsupervisedDomainAdaptation2015}
Tzu Ming~Harry Hsu, Wei~Yu Chen, Cheng-An Hou, Yao-Hung~Hubert Tsai, Yi-Ren
  Yeh, and Yu-Chiang~Frank Wang.
\newblock Unsupervised {{Domain Adaptation}} with {{Imbalanced Cross-Domain
  Data}}.
\newblock In \emph{IEEE/CVF International Conference on Computer Vision}, pages
  4121--4129, 2015.

\bibitem[Huang et~al.(2016)Huang, Li, Loy, and Tang]{Huang_2016_CVPR}
Chen Huang, Yining Li, Chen~Change Loy, and Xiaoou Tang.
\newblock Learning deep representation for imbalanced classification.
\newblock In \emph{Proceedings of the IEEE Conference on Computer Vision and
  Pattern Recognition (CVPR)}, 2016.

\bibitem[Huang et~al.(2022)Huang, Lu, Lin, Xie, and Lin]{aqt}
Wei-Jie Huang, Yu-Lin Lu, Shih-Yao Lin, Yusheng Xie, and Yen-Yu Lin.
\newblock Aqt: Adversarial query transformers for domain adaptive object
  detection.
\newblock In \emph{International Joint Conference on Artificial Intelligence
  (IJCAI)}, 2022.

\bibitem[Inoue et~al.(2018)Inoue, Furuta, Yamasaki, and Aizawa]{clipart}
Naoto Inoue, Ryosuke Furuta, Toshihiko Yamasaki, and Kiyoharu Aizawa.
\newblock Cross-domain weakly-supervised object detection through progressive
  domain adaptation.
\newblock In \emph{IEEE/CVF Conference on Computer Vision and Pattern
  Recognition}, 2018.

\bibitem[Jiang et~al.(2020)Jiang, Lao, Matwin, and
  Havaei]{jiangImplicitClassConditionedDomain2020}
Xiang Jiang, Qicheng Lao, Stan Matwin, and Mohammad Havaei.
\newblock Implicit {{Class-Conditioned Domain Alignment}} for {{Unsupervised
  Domain Adaptation}}.
\newblock In \emph{International Conference on Machine Learning}, pages
  4816--4827, 2020.

\bibitem[Johnson-Roberson et~al.(2017)Johnson-Roberson, Barto, Mehta, Sridhar,
  Rosaen, and Vasudevan]{sim10k}
Matthew Johnson-Roberson, Charles Barto, Rounak Mehta, Sharath~Nittur Sridhar,
  Karl Rosaen, and Ram Vasudevan.
\newblock Driving in the matrix: Can virtual worlds replace human-generated
  annotations for real world tasks?
\newblock In \emph{International Conference on Robotics and Automation}, pages
  746--753. IEEE, 2017.

\bibitem[Kar et~al.(2023)Kar, Chudasama, Onoe, and Wasnik]{Kar_2023_CVPR}
Purbayan Kar, Vishal Chudasama, Naoyuki Onoe, and Pankaj Wasnik.
\newblock Revisiting class imbalance for end-to-end semi-supervised object
  detection.
\newblock In \emph{IEEE/CVF Conference on Computer Vision and Pattern
  Recognition Workshops (CVPRW)}, pages 4570--4579, 2023.

\bibitem[Kennerley et~al.(2023)Kennerley, Wang, Veeravalli, and Tan]{2pcnet}
Mikhail Kennerley, Jian-Gang Wang, Bharadwaj Veeravalli, and Robby~T. Tan.
\newblock 2pcnet: Two-phase consistency training for day-to-night unsupervised
  domain adaptive object detection.
\newblock In \emph{IEEE/CVF Conference on Computer Vision and Pattern
  Recognition}, pages 11484--11493, 2023.

\bibitem[Kim et~al.(2019)Kim, Jeong, Kim, Choi, and Kim]{dm}
Taekyung Kim, Minki Jeong, Seunghyeon Kim, Seokeon Choi, and Changick Kim.
\newblock Diversify and match: A domain adaptive representation learning
  paradigm for object detection.
\newblock In \emph{IEEE/CVF Conference on Computer Vision and Pattern
  Recognition}, 2019.

\bibitem[Krishna et~al.(2023)Krishna, Ohashi, and Sinha]{mila}
Onkar Krishna, Hiroki Ohashi, and Saptarshi Sinha.
\newblock Mila: Memory-based instance-level adaptation for cross-domain object
  detection.
\newblock \emph{British Machine Vision Conference, (BMVC)}, 2023.

\bibitem[Li et~al.(2021)Li, Wu, Shrivastava, and Davis]{Li2021RethinkingPL}
Hengduo Li, Zuxuan Wu, Abhinav Shrivastava, and Larry~S. Davis.
\newblock Rethinking pseudo labels for semi-supervised object detection.
\newblock In \emph{AAAI}, 2021.

\bibitem[Li et~al.(2022{\natexlab{a}})Li, Liu, and Yuan]{sigma}
Wuyang Li, Xinyu Liu, and Yixuan Yuan.
\newblock Sigma: Semantic-complete graph matching for domain adaptive object
  detection.
\newblock In \emph{IEEE/CVF Conference on Computer Vision and Pattern
  Recognition}, 2022{\natexlab{a}}.

\bibitem[Li et~al.(2022{\natexlab{b}})Li, Dai, Ma, Liu, Chen, Wu, He, Kitani,
  and Vajda]{at}
Yu-Jhe Li, Xiaoliang Dai, Chih-Yao Ma, Yen-Cheng Liu, Kan Chen, Bichen Wu,
  Zijian He, Kris Kitani, and Peter Vajda.
\newblock Cross-domain adaptive teacher for object detection.
\newblock In \emph{IEEE/CVF Conference on Computer Vision and Pattern
  Recognition}, pages 7571--7580, 2022{\natexlab{b}}.

\bibitem[Liu et~al.(2021)Liu, Ma, He, Kuo, Chen, Zhang, Wu, Kira, and
  Vajda]{unbiased}
Yen-Cheng Liu, Chih-Yao Ma, Zijian He, Chia-Wen Kuo, Kan Chen, Peizhao Zhang,
  Bichen Wu, Zsolt Kira, and Peter Vajda.
\newblock Unbiased teacher for semi-supervised object detection.
\newblock In \emph{International Conference on Learning Representations}, 2021.

\bibitem[Oksuz et~al.(2020)Oksuz, Cam, Kalkan, and Akbas]{imbalance}
Kemal Oksuz, Baris~Can Cam, Sinan Kalkan, and Emre Akbas.
\newblock {Imbalance Problems in Object Detection: A Review}.
\newblock \emph{IEEE Transactions on Pattern Analysis and Machine
  Intelligence}, pages 1--1, 2020.

\bibitem[Ren et~al.(2015)Ren, He, Girshick, and Sun]{frcnn}
Shaoqing Ren, Kaiming He, Ross Girshick, and Jian Sun.
\newblock Faster r-cnn: Towards real-time object detection with region proposal
  networks.
\newblock In \emph{Advances in Neural Information Processing Systems}, page
  91–99, 2015.

\bibitem[Saito et~al.(2019)Saito, Ushiku, Harada, and Saenko]{strongweak}
Kuniaki Saito, Yoshitaka Ushiku, Tatsuya Harada, and Kate Saenko.
\newblock Strong-weak distribution alignment for adaptive object detection.
\newblock In \emph{IEEE/CVF Conference on Computer Vision and Pattern
  Recognition}, 2019.

\bibitem[Sakaridis et~al.(2018)Sakaridis, Dai, and Van~Gool]{foggycityscapes}
Christos Sakaridis, Dengxin Dai, and Luc Van~Gool.
\newblock Semantic foggy scene understanding with synthetic data.
\newblock \emph{International Journal of Computer Vision}, 126\penalty0
  (9):\penalty0 973--992, 2018.

\bibitem[Simonyan and Zisserman(2015)]{vgg}
Karen Simonyan and Andrew Zisserman.
\newblock Very deep convolutional networks for large-scale image recognition.
\newblock pages 1--14. Computational and Biological Learning Society, 2015.

\bibitem[Tanwisuth et~al.(2021)Tanwisuth, Fan, Zheng, Zhang, Zhang, Chen, and
  Zhou]{tanwisuthPrototypeOrientedFrameworkUnsupervised2021}
Korawat Tanwisuth, Xinjie Fan, Huangjie Zheng, Shujian Zhang, Hao Zhang, Bo
  Chen, and Mingyuan Zhou.
\newblock A {{Prototype-Oriented Framework}} for {{Unsupervised Domain
  Adaptation}}.
\newblock In \emph{Advances in Neural Information Processing Systems}, pages
  17194--17208, 2021.

\bibitem[Tarvainen and Valpola(2017)]{meanteacher}
Antti Tarvainen and Harri Valpola.
\newblock Mean teachers are better role models: Weight-averaged consistency
  targets improve semi-supervised deep learning results.
\newblock In \emph{Advances in Neural Information Processing Systems}, page
  1195–1204, 2017.

\bibitem[Tian et~al.(2019)Tian, Shen, Chen, and He]{fcos}
Zhi Tian, Chunhua Shen, Hao Chen, and Tong He.
\newblock Fcos: Fully convolutional one-stage object detection.
\newblock In \emph{IEEE/CVF International Conference on Computer Vision}, 2019.

\bibitem[Vs et~al.(2021)Vs, Gupta, Oza, Sindagi, and Patel]{mega}
Vibashan Vs, Vikram Gupta, Poojan Oza, Vishwanath~A Sindagi, and Vishal~M
  Patel.
\newblock Mega-cda: Memory guided attention for category-aware unsupervised
  domain adaptive object detection.
\newblock In \emph{IEEE/CVF Conference on Computer Vision and Pattern
  Recognition}, pages 4516--4526, 2021.

\bibitem[Wang et~al.(2017)Wang, Ramanan, and Hebert]{NIPS2017_147ebe63}
Yu-Xiong Wang, Deva Ramanan, and Martial Hebert.
\newblock Learning to model the tail.
\newblock In \emph{Advances in Neural Information Processing Systems}. Curran
  Associates, Inc., 2017.

\bibitem[Wu et~al.(2019)Wu, Kirillov, Massa, Lo, and Girshick]{detectron2}
Yuxin Wu, Alexander Kirillov, Francisco Massa, Wan-Yen Lo, and Ross Girshick.
\newblock Detectron2.
\newblock \url{https://github.com/facebookresearch/detectron2}, 2019.

\bibitem[Xu et~al.(2020{\natexlab{a}})Xu, Zhao, Jin, and
  Wei]{Xu2020ExploringCR}
Chang-Dong Xu, Xingjie Zhao, Xin Jin, and Xiu-Shen Wei.
\newblock Exploring categorical regularization for domain adaptive object
  detection.
\newblock \emph{IEEE/CVF Conference on Computer Vision and Pattern
  Recognition}, pages 11721--11730, 2020{\natexlab{a}}.

\bibitem[Xu et~al.(2020{\natexlab{b}})Xu, Wang, Ni, Tian, and Zhang]{gpa}
Minghao Xu, Hang Wang, Bingbing Ni, Qi Tian, and Wenjun Zhang.
\newblock Cross-domain detection via graph-induced prototype alignment.
\newblock In \emph{IEEE/CVF Conference on Computer Vision and Pattern
  Recognition}, 2020{\natexlab{b}}.

\bibitem[Yang and Soatto(2020)]{fda}
Yanchao Yang and Stefano Soatto.
\newblock {FDA}: Fourier domain adaptation for semantic segmentation.
\newblock In \emph{IEEE/CVF Conference on Computer Vision and Pattern
  Recognition}, pages 4084--4094, 2020.

\bibitem[Yoo et~al.(2022)Yoo, Chung, and Kwak]{oada}
Jayeon Yoo, Inseop Chung, and Nojun Kwak.
\newblock Unsupervised domain adaptation for one-stage object detector using
  offsets to bounding box.
\newblock In \emph{European Conference on Computer Vision}, pages 691--708.
  Springer, 2022.

\bibitem[Zhang et~al.(2022)Zhang, Pan, and Wang]{Zhang_Pan_Wang_2022}
Fangyuan Zhang, Tianxiang Pan, and Bin Wang.
\newblock Semi-supervised object detection with adaptive class-rebalancing
  self-training.
\newblock \emph{AAAI}, 36\penalty0 (3):\penalty0 3252--3261, 2022.

\bibitem[Zhang et~al.(2018)Zhang, Cisse, Dauphin, and Lopez-Paz]{mixup}
Hongyi Zhang, Moustapha Cisse, Yann~N. Dauphin, and David Lopez-Paz.
\newblock mixup: Beyond empirical risk minimization.
\newblock \emph{International Conference on Learning Representations}, 2018.

\bibitem[Zhao and Wang(2022)]{tia}
Liang Zhao and Limin Wang.
\newblock Task-specific inconsistency alignment for domain adaptive object
  detection.
\newblock In \emph{IEEE/CVF Conference on Computer Vision and Pattern
  Recognition}, 2022.

\bibitem[Zhao et~al.(2023)Zhao, Wei, Chen, Li, Yang, Peng, and Liu]{mrt}
Zijing Zhao, Sitong Wei, Qingchao Chen, Dehui Li, Yifan Yang, Yuxin Peng, and
  Yang Liu.
\newblock Masked retraining teacher-student framework for domain adaptive
  object detection.
\newblock In \emph{Proceedings of the IEEE/CVF International Conference on
  Computer Vision (ICCV)}, pages 19039--19049, 2023.

\bibitem[Zheng et~al.(2020)Zheng, Wu, Han, and Shi]{forkgan}
Ziqiang Zheng, Yang Wu, Xinran~Nicole Han, and Jianbo Shi.
\newblock Forkgan: Seeing into the rainy night.
\newblock In \emph{European Conference on Computer Vision}, 2020.

\bibitem[Zhu et~al.(2017)Zhu, Park, Isola, and Efros]{cyclegan}
Jun-Yan Zhu, Taesung Park, Phillip Isola, and Alexei~A. Efros.
\newblock Unpaired image-to-image translation using cycle-consistent
  adversarial networks.
\newblock In \emph{IEEE/CVF International Conference on Computer Vision}, pages
  2242--2251, 2017.

\bibitem[Zhu et~al.(2021)Zhu, Su, Lu, Li, Wang, and Dai]{defdetr}
Xizhou Zhu, Weijie Su, Lewei Lu, Bin Li, Xiaogang Wang, and Jifeng Dai.
\newblock Deformable {\{}detr{\}}: Deformable transformers for end-to-end
  object detection.
\newblock In \emph{International Conference on Learning Representations}, 2021.

\end{thebibliography}
